# A Simulation Study of Functional Electrical Stimulation for An Upper Limb Rehabilitation Robot using Iterative Learning Control (ILC) and Linear models.

Boluwatife E. FAREMI [a, d, *], Kayode P. AYODELE [a, *], Abimbola M. JUBRIL [a, *], Afeez A. FAKUNLE [a], Mathew O.B. OLAOGUN [b], Micheal B. FAWALE [c], Morenikeji A. KOMOLAFE [c]

**Abstract**

A proportional iterative learning control (P-ILC) for linear models of an existing hybrid stroke rehabilitation scheme is implemented for elbow extension/flexion during a rehabilitative task. Owing to transient error growth problem of P-ILC, a learning derivative constraint controller was included to ensure that the controlled system does not exceed a predefined velocity limit at every trial. To achieve this, linear transfer function models of the robot end-effector interaction with a stroke subject (plant) and muscle response to stimulation controllers were developed. A straight-line point-point trajectory of 0 - 0.3 m range served as the reference task space trajectory for the plant, feedforward, and feedback stimulation controllers. At each trial, a SAT-based bounded error derivative ILC algorithm served as the learning constraint controller. Three control configurations were developed and simulated. The system's performance was evaluated using the root means square error (RMSE) and normalized RMSE. At different ILC gains over 16 iterations, a displacement error of 0.0060 m was obtained when control configurations were combined.

## 1. Introduction

Stroke is a neurological loss that arises from death of brain cells due to obstruction of proper flow of blood or injuries leading to blood spillage within the cerebrum [1-2]. According to [3], stroke rehabilitation is the restoration of brain tissue and networks via relearning and compensation for lost functional abilities. The process involves plastic re-organization of entire regions and brain pathways for the generation of commands to the same muscles that produced unique motor actions and patterns before the occurrence of stroke [4].

It has been observed that patient experience faster recovery and improved outcome when highly intense task are adopted during rehabilitation sessions [5]. However, the sessions are time consuming and energy intensive, as a result, rehabilitation robots are incorporated into training sessions to invoke plastic changes in the appropriate neuronal populations and connections while ensuring the task offered to patient's are challenging [6].

Hybrid combination of FES and robot-assisted paradigm is considered for upper-limb rehabilitation because of the multiple benefits of intense training via robots, excitation of muscular pathways for movement and spasticity reduction via FES [7][8]. Notwithstanding, most FES schemes are open-loop systems that apply predetermined signals without appropriate considerations of patient improvements. [9]. This approach induces muscle fatigue and makes most patient's FES dependent after therapy [10].

Recent applications of feedback and feedforward controllers into FES design have been reported [11].

a Department of Electronic and Electrical Engineering, Obafemi Awolowo University, Ile-Ife, Osun 220005, Nigeria.
b Faculty of Medical Rehabilitation, University of Medical Sciences, Ondo City, Nigeria.
c Department of Medicine, Obafemi Awolowo University, Ile-Ife, Osun 220005, Nigeria.
d School of Computing Sciences and Computer Engineering, The University of Southern Mississippi, Hattiesburg, MS 39401, USA.

The feedback controller ensures stabilization of muscle response during tracking tasks, while the feedforward controller calculates a feedforward corrective signal for satisfactory tracking during next task [12]. To obtain better performance using these controllers, models of the muscle, arm and the robot are reported to be vital [9].

A model based ILC design that combined feedforward and feedback controllers with FES is presented in [13]. The system varies the intensity of FES delivered to patients using an exoskeleton robot. The implemented hybrid scheme was reported to improve patient's performance while undergoing a task.

Another hybrid active-based planar robotic system with FES was presented in [14]. The study adopted a human arm model, feedback controller, Phase-lead and Adjoint ILC control scheme for adjusting the intensity of FES. Furthermore, [15] designed an electrode-array based hybrid rehabilitation system. In the study, a joint robot-musculoskeletal model was considered in designing a closed-loop FES system. The adopted control involves an iterative learning controller for controlling the pulse-width based on task improvements. Additionally, [16] developed a hybrid system that combined musculoskeletal model and intentional effort to mediate FES. The study combined iterative learning control, EMG and FES to improve patient outcome during rehabilitation using a planar robot. Experimental results showed significant tracking error reduction compared to traditional controllers.

However, one major challenge with ILC method is the transient learning error growth problem where the error increases at the onset of the learning process. This may be intolerable and harmful to the patient. This also causes excessive control inputs to actuators (biological muscles) which have limits to minimize the error, improve performance and attain a desired convergence. [17 - 19]. It should be noted that most implementation of ILC in a hybrid scheme lack a form of input or output constraint [18]. Another consequence of this initial excessive control signal is a high velocity. This may also lead to injury to the already weak nerves.

Therefore, a form of constraint is required in hybrid schemes to avoid excessive control efforts to the feeble muscle system of impaired patients. Additionally, it is also deemed fit to avoid exceeding a patient-identified or specific velocity.

The contribution of the present paper is the development of two algorithms that corrects patient upper-limb movement error from trial to trial via a proportional ILC (P-ILC) and a new learning velocity constraint algorithm that ensures patient-identified velocity isn't exceeded based on a modified bounded error algorithm + SAT function + derivative type ILC (D-ILC). This paper is organized as follows: In section 2, interaction dynamics of the robot system. Section 3 describes the approximate linear models and control schemes. Section 4, 5 and 6 presents the result, discussion and conclusion, respectively.

## 2. THEORY

### 2.1 INTERACTION DYNAMICS OF ROBOT.

The hybrid rehabilitation robot system adopted in this study is the platform for upper limb robot (PULSR) workstation developed at Obafemi Awolowo University, Nigeria for flexion and extension about the elbow joint on a planar surface (2D cartesian plane). As shown in Fig. 1, it consists of five links with four of them set up as a standard parallelogram while the fifth link serves as a mounting region for the stroke patient forearm. To capture the dynamics that exist between the patient and the robot at the point of contact B (end-effector), a second-order impedance model given in Equation (1) was used [20]. This scheme offers a virtual means of creating a challenging task and controlling the contact force felt by the patient through programmable stiffness and dampness matrices [21].

$$f_r = K_{M_x}\ddot{\tilde{x}}_r - K_{B_x}\dot{\tilde{x}}_r - K_{K_x}\tilde{x}_r \qquad (1)$$

where $\tilde{x}_r = x_{ref} - x_r$, $\theta_r = [\theta_1, \theta_2]^T$, $x_r = k(\theta_r)$, $\dot{x}_r = J_r(\theta_r)\dot{\theta}_r$, $\ddot{x}_r = J_r(\theta_r)\ddot{\theta}_r + \dot{J}_r(\theta_r, \dot{\theta}_r)\dot{\theta}_r$.

To prescribe the force effect that the vector of robot joint $\theta_r$ should apply at the end-effector on the forearm and elbow joint of the subject, the gains $K_{M_x}$, $K_{B_x}$, $K_{K_x}$ were dynamically computed over an experimental procedure.

## 3. METHODS

The objective of this section is to discuss an adopted model structure from [14], that will capture the combined interaction of the robot and the elbow region to be rehabilitated. In the present work, more emphasis is placed on stimulating the muscles to drive task completion rather than the robot. Accordingly, the robot end-effector dynamics are tuned to produce an effect of shifting a mass during tracking task. Additionally, the linearized muscle model utilized to drive the forearm is described. Afterwards, control strategies involved in developing appropriate stimulation schemes are implemented as shown in Fig. 2.

### 3.1 Approximate human and robot model

In the present section, an approximate model is adopted because it offers a means to rapidly iterate and evaluate if control algorithms to be developed will be beneficial to subjects before lengthy rehabilitation process. A single input and single output (SISO), linear, time-invariant transfer function that gives a relationship between output forearm angle and linear activated muscular contraction at the input is chosen. According to [14], Equation 2 incorporates a linear-approximate model of a stroke patient interacting with a 5-link manipulator robot like PULSR.

$$\frac{\vartheta_f}{T_\beta} = \frac{1}{s\left((b_{a_3}+K_{M_2})s+K_{B_2}\right)} \quad (2)$$

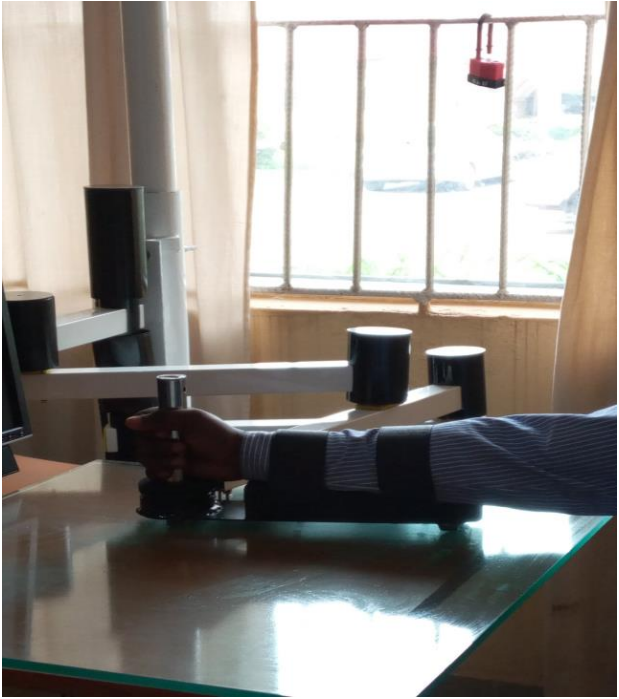

Figure 1: An image of the PULSR rehabilitation robot.

where,

$\vartheta_f$ = Elbow joint displacement due to muscle activation.
$T_\beta$ = Muscle activation due to applied stimulation.
$K_{B_2}$, $K_{M_2}$ = robot end-effector gains for defining the virtual effect of a moving mass.

$$b_{a_3} = m_f l_{f_1}^2 + I_f + I_e \left(\frac{s_\gamma}{1-c_f^2 c_\gamma^2}\right)^2 \quad (3)$$

where,

$c_{(.)}, s_{(.)}$ = denote cosine and sine of elbow angles.

$m_f$ = mass of forearm.

$I_f = l_{f_1} + l_{f_2}$ = fore-arm length of the subject.
$I_e$ = Moment of the forearm.
$\gamma$ = Static angle of elevation.
$\frac{s_\gamma}{1-c_f^2 c_\gamma^2}$ = defines the region in the elbow, where stimulus is applied.

Table 1: Parameters describing the Arm dynamics and Task-plane trajectory

| Description | Notation | Data |
|---|---|---|
| Mass of forearm | $m_f$ | 0.84kg |
| Length of forearm from the olecranon joint to the forearm center of gravity | $l_{f_1}$ | 0.203m |
| Length of the forearm from the center of gravity to thumb web | $l_{f_2}$ | 0.203m |
| Length of the upperarm from the shoulder joint to the upperarm center of gravity | $l_{u_1}$ | 0.154m |
| Length of the upperarm from the center of gravity to olecranon region. | $l_{u_2}$ | 0.154m |
| Inertia of forearm | $I_f$ | 0.12 |
| Moment of forearm | $I_e$ | 0.15 |
| Constrained Forearm Elevation angle | $\gamma$ | 1.0472rad |
| Maximum Length of reach (MALOR) from glenohumeral joint base | $r_2$ | 0.3m |
| Minimum length of reach (MILOR) from glenohumeral joint base | $r_1$ | 0.1m |
| Task space orientation angle | $\phi$ | 0.6109rad |
| Horizontal distance from Subject's body | $d1$ | 0.2m |

Substituting measured anthropometric data obtained from [20] in Table 1, the resulting transfer function of the patient arm-robot model (plant) is given as

$$T\_ar(s) = \frac{1}{0.5571s^2 + 5.78s} \quad (4)$$

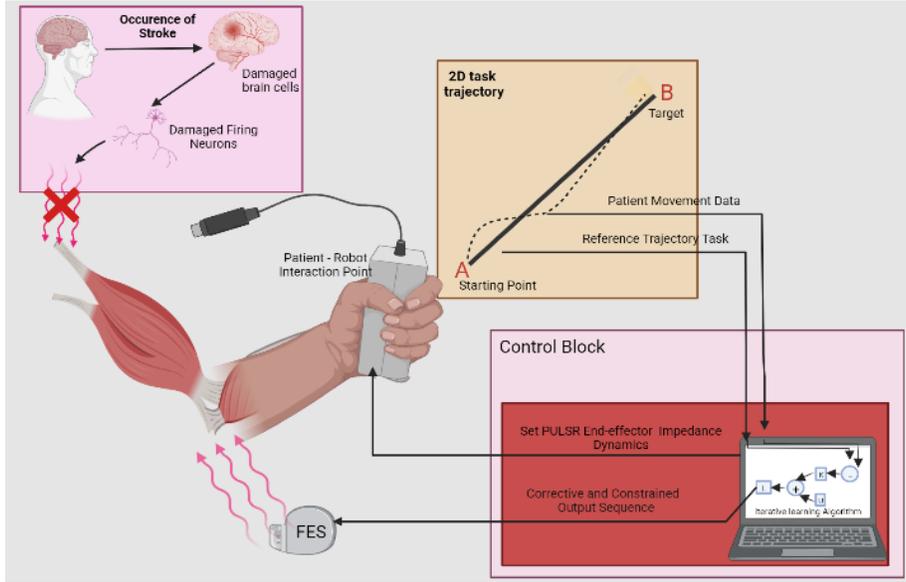

Figure 2: Diagram showing the upper limb interaction + PULSR end-effector + developed control schemes

## 3.2 Approximate muscle model

The Hill type model described by [22] has been selected as the model for representing the muscle (biological actuator) in this work. Equation 5 describes the torque generated by the muscle as a function of stimulation pulse-width ($W_{PD}$), passive multiplicative effect of forearm length ($F_{mp}$), active multiplicative effect of forearm displacement and velocity ($F_{ma}$).

$$T_\beta = g(W_{PD}) \times F_{ma}(\dot{\vartheta}_f) + F_{mp}(l_f) \quad (5)$$

where,

$W_{PD}$ = input stimulation.

Hammerstein model = $g(W_{PD}) = H_{irc} \times H_{lad}$. (5.1)

From [23], the isometric recruitment $H_{irc}$ and linear activation dynamics $H_{lad}$ can be modelled using Equations (5.2) and (5.3)

$$H_{irc} = f(u) = a_1 \cdot \frac{e^{a_2 \cdot u} - 1}{e^{a_2 \cdot u} + a_3} \quad (5.2)$$

$$H_{lad} = \frac{w_n^2}{s^2 + 2sw_n + w_n^2} \quad (5.3)$$

where,

$a_1, a_2$ and $a_3$ = static gain parameters.

$W_{PD}$ or $u$ = stimulation input passed into the hammerstein model.

$w_n$ = encapsulation of muscle dampness.

From [24], $F_{ma}$ and $F_{mp}$ can be modelled as Equation 5.4 and Equation 5.5.

$$F_{ma}(\dot{\vartheta}_f) = 0.54 \tan^{-1}(5.69\dot{\vartheta}_f + 0.51) + 0.745 \quad (5.4)$$

$$F_{mp}(l_f) = e^{-\left(\frac{\bar{l}_f - 1}{\varepsilon}\right)^2} \quad (5.5)$$

where,

$\dot{\vartheta}_f$ = normalized velocity = $\frac{\vartheta_f}{\vartheta_{max}}$

$\bar{l}_f$ = normalized forearm length = $\frac{l_f}{l_{max}}$

$\varepsilon$ = shape-factor of force-length relationship.

To obtain a linear muscle model, an inverse dynamics linearization was applied to Equation (5), using Equation (5.6). Fig. 3, depicts the linearization approach.

$$T_\beta = \left(\left(\left(\left(W_{PD} + F_{mp}\right)\left(H_{irc}^{-1}(H_{LAD} \times H_{irc})\right)\right) \times F_{ma}^{-1}\right) \times F_{ma}\right) - F_{mp} \quad (5.6)$$

The resulting structure leaves only the $H_{lad}$ as function of input stimulus in Equation (5.7). Substituting the desired damping into (5.3), the muscle model is given in (5.8)

$$\frac{T_\beta}{W_{PD}} = \frac{w_n^2}{s^2 + 2sw_n + w_n^2} \quad (5.7)$$

$$T\_mus(s) = \frac{7.129}{s^2 + 5.34s + 7.129}$$

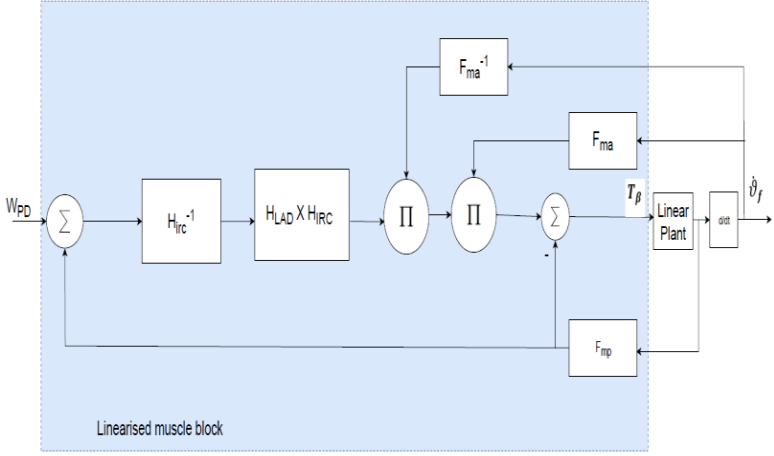

Figure 3: Inverse dynamics linearization of muscle model

### 3.3 Controller Designs

The aim of this section is to design controllers that supply appropriate control signals needed for stimulating the muscle system towards displacement of the patient arm-robot model and to ensure the supplied control signals do not exceed patient-specific velocity.

#### 3.3.1 Feedback controller

Generally, the feedback controller helps to improve the transient response of the unresponsive flaccid muscle of impaired subjects during rehabilitative task. In the present work, purpose of this controller is to apply necessary control demand that ensures the linear muscle dynamics produces a stable response for plant (arm-robot model) tracking. As a result, an approximate form of a PD controller known as phase-lead compensator was selected to improve the transient response of the muscle. [25] outlines the generic type of lead compensator as

$$\frac{W_{PD}}{U} = \frac{K_P}{1} \frac{\tau s+1}{\varpi \tau s+1} \tag{6}$$

With $\tau = \frac{K_D}{K_P}$ and $0 < \varpi < 1$ \hspace{1em} (7)

The expression in (6) can be re-written as

$$\frac{W_{PD}}{U} = \frac{(\frac{K_D}{K_P}s+1)K_P}{\varpi \frac{K_D}{K_P}s+1} \tag{8}$$

As $\varpi \frac{K_D}{K_P} \to 0$, the numerator approximates a typical PD controller. The gains $K_D, K_p$ were manually tuned over statistically observed value range.

The resulting negative feedback transfer function of the entire system is given as

$$H(s) = \frac{14.26s+71.29}{0.005571s^5+0.6447s^4+9.103s^3+35.25s^2+55.46s+71.29} \tag{9}$$

The closed loop-system performance to step input and the controlled muscle activation was evaluated over 10 seconds and observed responses are depicted in Fig. 4 and 5 respectively.

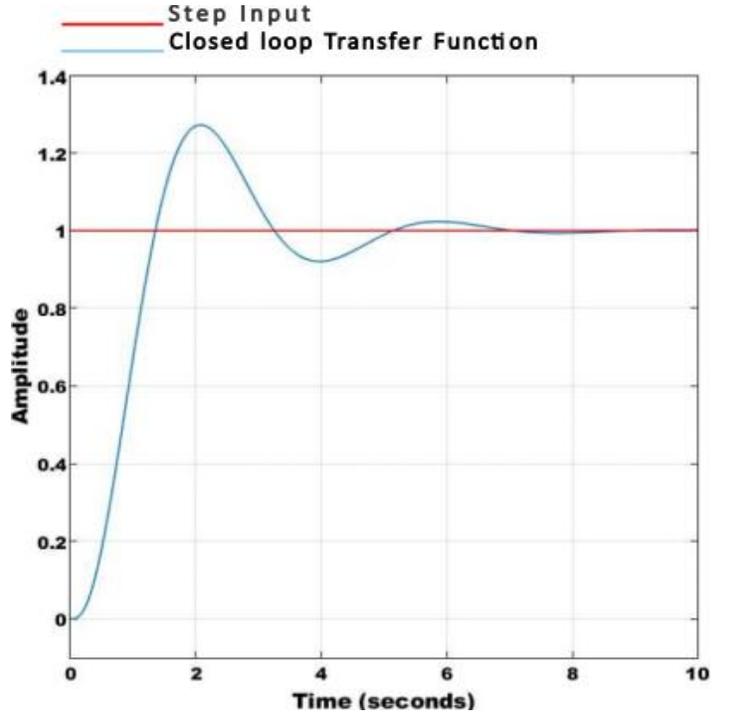

Figure 4: System response to step input.

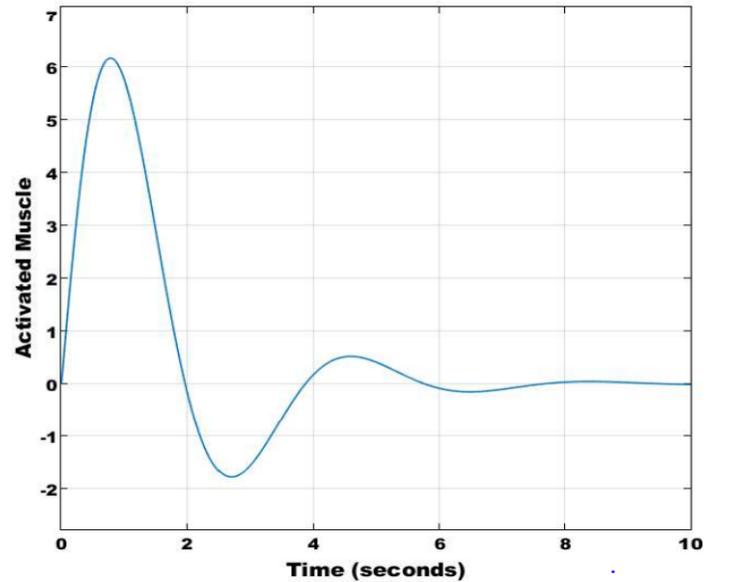

Figure 5: Muscular Response during closed loop.

### 3.3.2 Feedforward controller development.

In this section, iterative learning control is used to attain the goal of improving tracking performance of the plant and reduce tracking error from attempt to attempt.

In the present framework, the P-ILC was selected as the controller type because of its simplicity and structure. Since ILC is a two-dimensional process [32] (i.e., It is indexed over an iteration domain and time domain), the following considerations were made to ascertain good performance.

- Definition of finite time interval for simulation/rehabilitation session.
- The system to be controlled.
- The bandwidth of the closed loop system must be known.
- A Q-filter that removes input signals that are above system's bandwidth at every iteration if noises exist in the feedforward input.
- Selection of a learning gain.
- Definition of the desired reference task plane trajectory to be tracked.

The control efforts that will be stored at each trial are:

1) Previous input signal.
2) Output displacement signal.
3) Error signal.

According to [26], an ILC learning algorithm for linear-time invariant system is given as

$$U_{k+1}(t) = Q(U_k(t) + Le_K(t))$$
$$e_k(t) = Y_r(t) - Y_k(t) \quad (10)$$

where, $U_k(t)$ is the ILC output over the present iteration index, $e_k(t)$ is the Error at present iteration index, $Y_r(t)$ = reference trajectory, $U_{k+1}(t)$ = Next trial feedforward input, $L$ = Proportional learning gain and Q = Filter.

However, [27] disclosed that, attaining monotonic convergence with Equation (10) is dicey because of the absence of auto-tuning guidelines. To solve this problem, the study suggested including a lowpass filter in order to disable learning at higher frequencies. Furthermore, the study claimed that if the iterations are sufficiently short, desired convergence could be found. To affirm this claim, a reset-test-run method was disclosed by [28] for conducting or selecting learning gains that provide better convergence over desired iterations [27 -28].

Accordingly, the bandwidth of the closed loop system was first obtained as Equation (11) to (12) and shown in Fig. 6

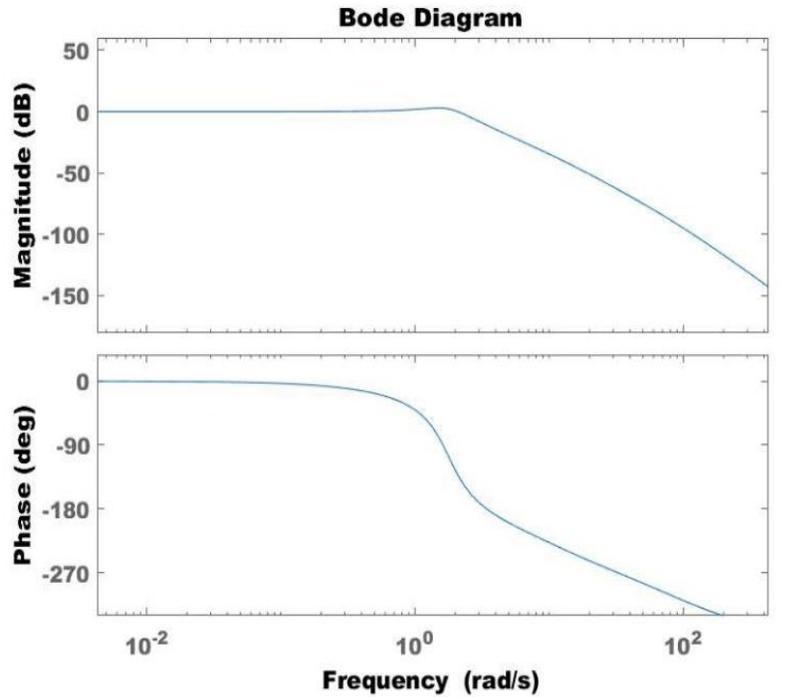

Figure 6: Bode plot of the entire closed-loop system

$$w(rad/s) = 2\pi f, f(Hz) = \frac{w}{2\pi} \quad (11)$$

$$f(Hz) = \frac{2.465}{2\pi} = 0.392 \approx 0.40 Hz \quad (12)$$

Thereafter, a first-order low-pass filter transfer function structure in Equation (13) was designed for eliminating signals that will degrade the closed loop system performance in which ILC will act on.

$$Q(s) = \frac{\omega_c}{s+\omega_c} = \frac{1}{1+\frac{s}{\omega_c}} \quad (13)$$

where $\omega_c = 2\pi \cdot 0.40$

The choice of Learning Filter was selected based on several simulations that showed monotonically decreasing plant tracking error and convergence.

$$L \in [0.1, 1] \quad (14)$$

### 3.3.3 Development of learning velocity constraint controller.

The choice of learning constrain-velocity controller is a D-ILC based on bounded error algorithm by [29] and soft-saturation constraint by [17]. Rooted on the evidence from literature that ILC will increases its control efforts for error minimization, it's desired to have a control scheme that ensures the transient error growth problem of P-ILC doesn't cause the feedback controller to apply control signals that will violate the identified maximum velocity for the plant. Hence, the aim of the controller is to learn a suitable stimulation control signal and apply a saturation constraint on the feedback controller based on the computed end-effector velocity error between the identified maximum velocity and velocity obtained from trial to trial.

Problem Formulation,

i. Let the derivative learning controller be defined as

$$V_{k+1} = V_k + \psi \dot{e}_k \quad (15)$$
where,

$i$ = sample instance

$V_k$ = initial constrain on the PD controller's output at the first iteration.

$\dot{e}_k = \dot{r}_h - \dot{r}_i$ = difference between the identified maximum resultant velocity of the plant in absence of P-ILC and the resultant velocity during P-ILC based trials.

$\psi$ = slow learning derivative gain.

To avoid transient error growth in D-ILC controller, a slow learning rate $\psi$ was selected out of the methods given by [30] for solving transient error growth.

Moreover, to obtain $\dot{r}_h$ for every patient to be rehabilitated, the derivative of time step functions described by [31] in Fig. 7 was used to obtain $\dot{\vartheta}_f$. Next, the component of vector of end-effector velocities were obtained using

$$\dot{x}_h = J_h(\vartheta_f) \begin{bmatrix} \hat{k}(\dot{\vartheta}_f) \\ \dot{\vartheta}_f \end{bmatrix} \quad (16)$$

where,

$J_h(\vartheta_f)$ = Jacobian of human joint.

$\vartheta_f, \dot{\vartheta}_f$ = joint displacement signal and derived joint velocity.

$\hat{k}(\dot{\vartheta}_f) = \dot{\vartheta}_u$ = Approximate velocity component in shoulder joint.

$\dot{x}_h = \begin{bmatrix} \dot{x} \\ \dot{y} \end{bmatrix}$ = component of velocities along x and y axis on the task plane.

$\dot{r}_h$ = resultant velocity of velocity components $\dot{x}_h$ along the trajectory.

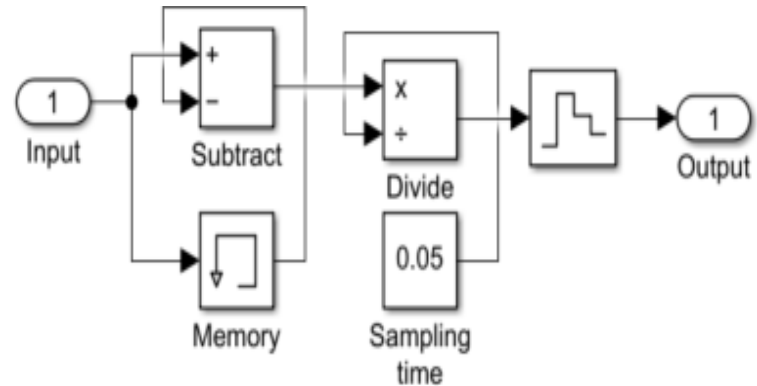

Figure 7: Derivative of elbow joint to obtain elbow velocity.

ii. To update the constrain input on the PD controller's output at the next trial via $V_{k+1}$, the bounded error inequality given by [29] in Equation 17 must be satisfied by $\dot{e}_k$.

$$\varepsilon_k = min(min(\dot{r}_h^{max} - \dot{r}_i), min(\dot{r}_i - \dot{r}_h^{min})) > 0 \quad (17)$$

where,

$\dot{r}_h^{min}$ = represents resultant velocity and no movement.

$\dot{r}_h{}^{max}$ = identified maximum resultant velocity an impaired subject can undertake.

$\dot{r}_i = n \times 1$ resultant velocity at an iteration.

ii. At any instance where $\varepsilon_k$ is violated, the update law in Equation 18, is computed to ensure $\dot{r}_h{}^{max}$ is not violated at the next trial instead of Equation 15.

$$V_{k+1} = V_k - \psi \dot{e}_k \qquad (18)$$

where,

$V_{k+1} = n \times 1$ update velocity constraint to be imposed at next iteration.

iii. Next, is the implementation of a soft saturation constraint by [17] on PD controller output signals.

$$sat = \big(max(V_{k+1}), u_{pd}\big) \qquad (19)$$

where,

$u_{pd}$ = signals for muscle stimulation.
$max(V_{k+1})$ = maximum computed constrain signal on PD controller.

3.4 **Simulation Scenarios**

The models and control strategies discussed in this study were implemented in SIMULINK and MATLAB. At a simulation time of 10 seconds and a sampling time 0.05 seconds, the formulated control structure and all other building blocks were built and updated into SIMULINK. Furthermore, a variant of the line-line (i.e., straight line) trajectory task is used as the 2D surface trajectory.

Using equation 20-22, a straight-line trajectory of range 0 – 0.2828 m was designed as the 2D task-plane trajectory for the plant. Fig. 8 depicts how the trajectory will look on a 2D surface when projected.

$$M_{agmitude} = \sqrt{(x_2 - x_1)^2 + (y_2 - y_1)^2} \qquad (20)$$

$$r = \sqrt{d1^2 + L^2} \qquad (21)$$

$$\phi = tan^{-1}\frac{d1}{L} \qquad (22)$$

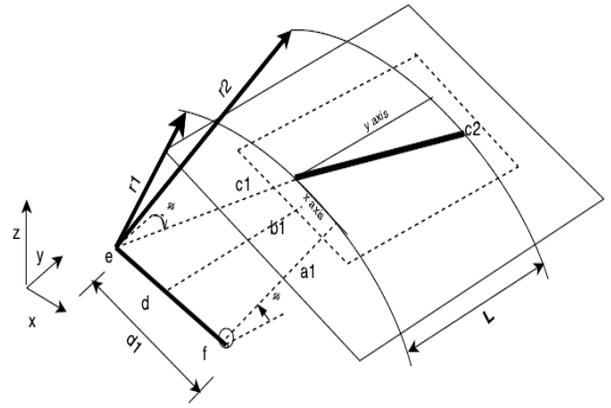

Figure 8: Side view of the straight-line trajectory at angle $\phi$ on the 2D surface

In the simulation procedure, the duty-cycle of a modelled FES block was modulated by the feedback controller signals. The amplitude and frequency of this block were kept constant at 5 mA and 50 Hz respectively. Three simulation instances were used to evaluate the efficacy of each controller influence on plant tracking performance.

First scenario examines the effect of PD controller on plant tracking performance. The second scenario examines the influence of the feedforward P-ILC controller when combined with existing feedback controller. The last scenario examines the effectiveness of the learning velocity constrain controller on the transient error growth problem of P-ILC on the PD, and the tracking performance of the preceding scenarios. Two metrics were then used to describe the performance of the control strategies. The normalized root means square error and the root mean square error

$$RMSE = \sqrt{\frac{1}{n}\sum_{k}^{n}(x_d - x_i)^2} \qquad (23)$$

$$NRMSE = \frac{\sqrt{\frac{1}{n}\sum_{k}^{n}(x_d - x_i)^2}}{x_{d_{max}} - x_{d_{min}}} \qquad (24)$$

## 4 RESULTS

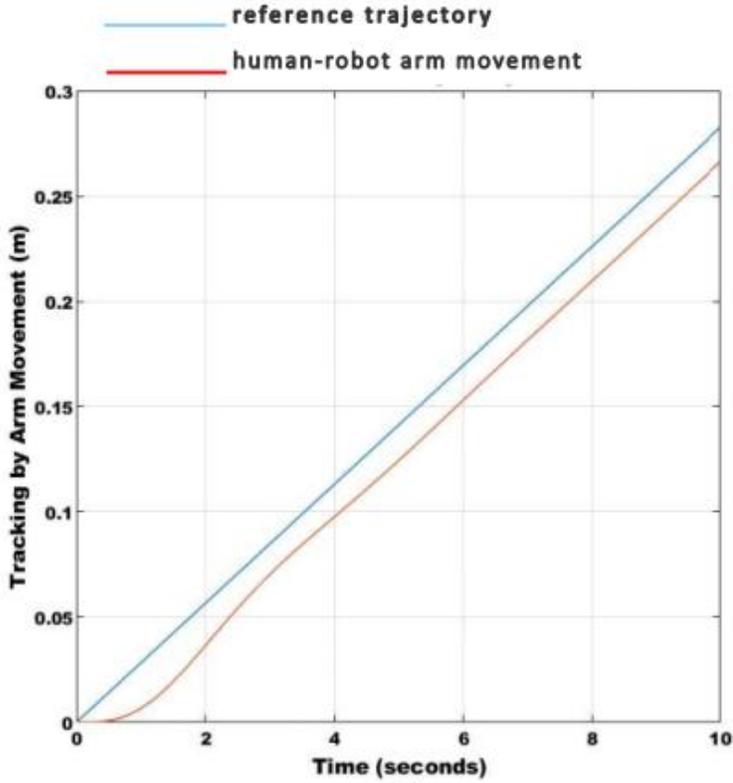

Figure 9: Plant tracking performance along 2d trajectory under PD controller only.

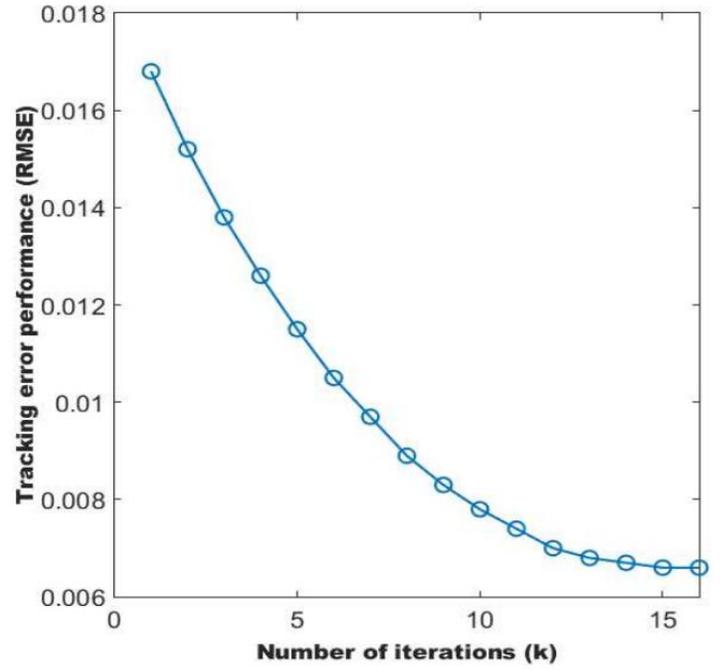

Figure 10: Influence of P-ILC on minimization of displacement errors over sixteen iterations.

Table (2): A table of non-decreasing displacement error when the feedback scheme is used as the only controller over the straight line trajectory in 2D space.

| No of iterations (k) | Constant PD controller displacement error in meters. RMSE(NRMSE) |
|---|---|
| 1 | 0.0168(0.0594) |
| 2 | 0.0168(0.0594) |
| 3 | 0.0168(0.0594) |
| 4 | 0.0168(0.0594) |
| 5 | 0.0168(0.0594) |
| 6 | 0.0168(0.0594) |
| 7 | 0.0168(0.0594) |
| 8 | 0.0168(0.0594) |
| 9 | 0.0168(0.0594) |
| 10 | 0.0168(0.0594) |

Table (3): Displacement error of plant in meters and percentage "NMSE(RMSE)" over the trajectory range at learning gains.

| No of iterations | L = 0.1 | L = 0.2 | L = 0.8 | L = 0.9 |
|---|---|---|---|---|
| 1 | 0.0168(0.0594) | 0.0168(0.0594) | 0.0168(0.0594) | 0.0168(0.0594) |
| 2 | 0.0152(0.0537) | 0.0137(0.0484) | 0.0064(0.0226) | 0.0060(0.0212) |
| 3 | 0.0138(0.0487) | 0.0113(0.0399) | - | - |
| 4 | 0.0126(0.04455) | 0.0095(0.0335) | - | - |
| 5 | 0.0115(0.4066) | 0.0081(0.0286) | - | - |
| 6 | 0.0105(0.0371) | 0.0072(0.0254) | - | - |
| 7 | 0.0097(0.0342) | 0.0067(0.0236) | - | - |
| 8 | 0.0089(0.0314) | 0.0066(0.0233) | - | - |
| 9 | 0.0083(0.0293) | - | - | - |
| 10 | 0.0078(0.0275) | - | - | - |
| 11 | 0.0074(0.0261) | - | - | - |
| 12 | 0.0070(0.0247) | - | - | - |
| 13 | 0.0068(0.0240) | - | - | - |
| 14 | 0.0067(0.0236) | - | - | - |
| 15 | 0.0066(0.0233) | - | - | - |
| 16 | 0.0066(0.0233) | - | - | - |

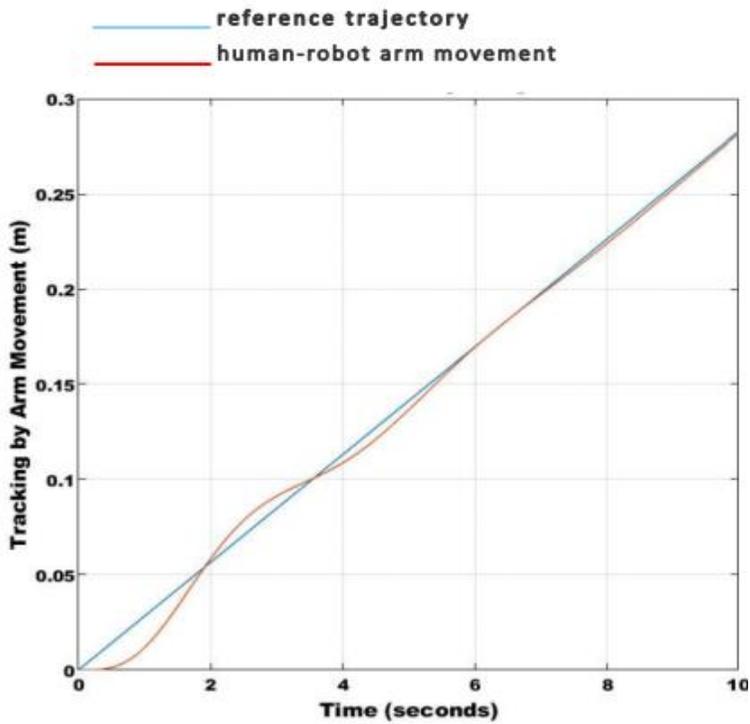

Figure 11: A plot of plant tracking performance and minimal displacement error along the 2D trajectory over sixteen iterations.

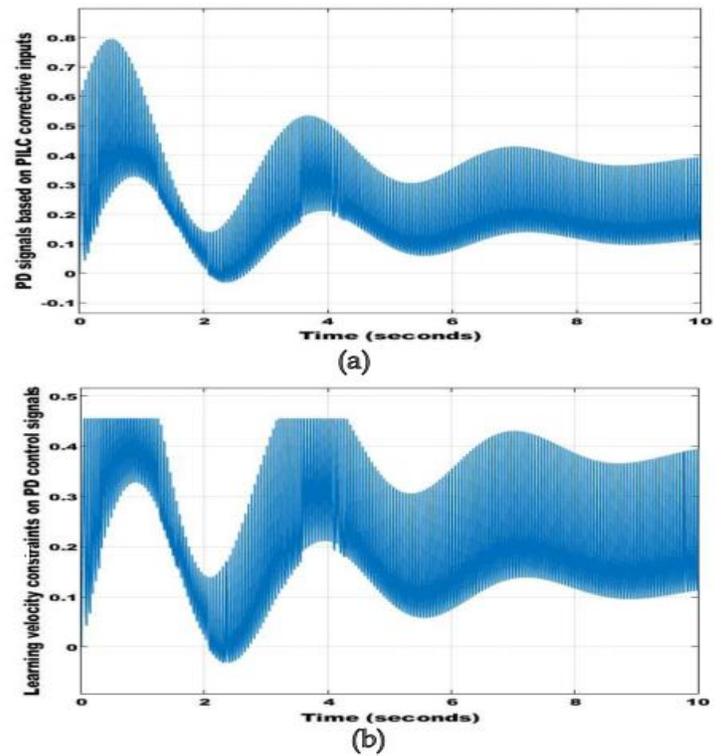

Figure 13: Increasing PD signals at the thirteen iteration and the effect of the D-ILC learning controller

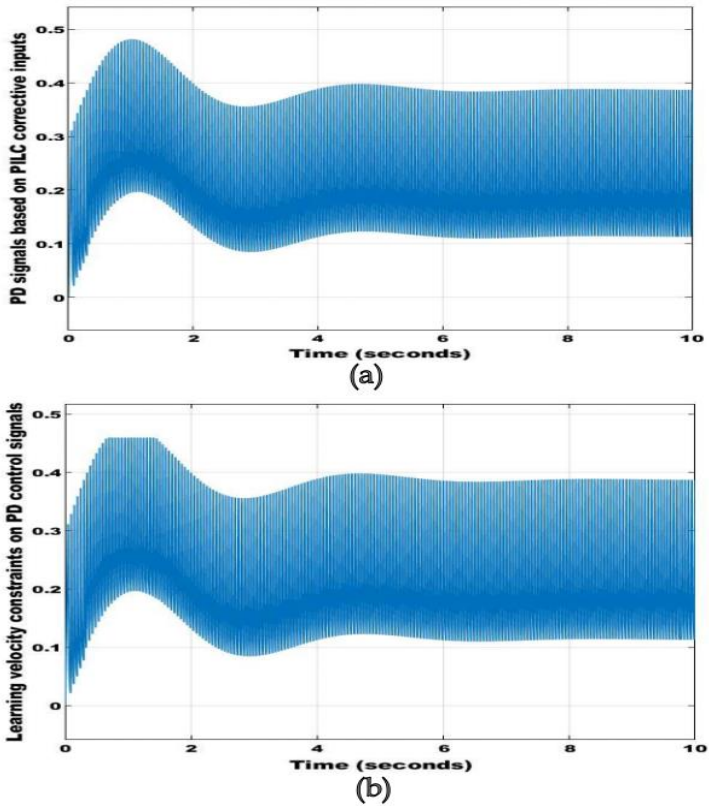

Figure 12: Growing and constrained PD signals at 1st iteration. (a) Increased PD signals due to P-ILC transient growth (B) Applied constraint on PD signals by the Learning D-ILC

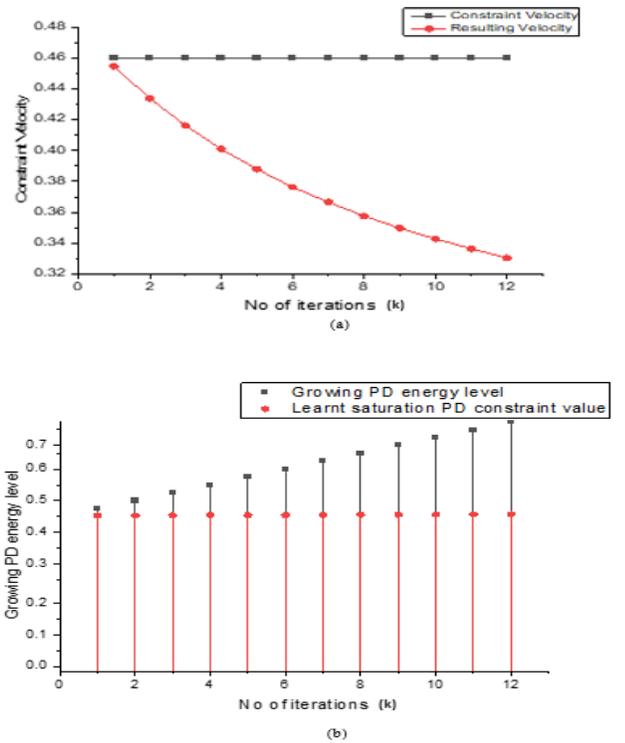

Figure 14: Plots corresponding to applied D-ILC effect on constraining velocity and growing PD control signal due to P-ILC (a) Plot of desired and resulting velocity (b) Plot of imposed D-ILC learning velocity constraint on growing PD controller signal.

Table (4): A table describing growing PD energy signals due to P-ILC, D-ILC applied learning constraint and resulting plant velocities obtained over thirteen trials

| No of iteration | Developed PD energy signals in absence of P-ILC corrective inputs | Developed PD energy signals due to P-ILC corrective inputs | Plant Specific velocity. (m/s) | Applied constraint PD energy signals by D-ILC | Resulting Velocity (m/s) | Next Constraint PD signals |
|---|---|---|---|---|---|---|
| (0) | 0.4615 | | 0.4791 | - | | 0.4603 |
| 1 | | 0.4814 | | 0.4603 | 0.4546 | 0.4605 |
| 2 | | 0.5047 | - | 0.4605 | 0.4338 | 0.4609 |
| 3 | | 0.5301 | - | 0.4609 | 0.4162 | 0.4614 |
| 4 | | 0.5574 | - | 0.4614 | 0.4011 | 0.4618 |
| 5 | | 0.5856 | - | 0.4618 | 0.3879 | 0.4622 |
| 6 | | 0.5856 | - | 0.4622 | 0.3763 | 0.4626 |
| 7 | | 0.6144 | - | 0.4626 | 0.3666 | 0.4629 |
| 8 | | 0.6434 | - | 0.4629 | 0.3576 | 0.4633 |
| 9 | | 0.7034 | - | 0.4633 | 0.3498 | 0.4637 |
| 10 | | 0.7341 | - | 0.4637 | 0.3428 | 0.4640 |
| 11 | | 0.7644 | - | 0.4640 | 0.3364 | 0.4644 |
| 12 | | 0.7953 | - | 0.4644 | 0.3304 | 0.4647 |

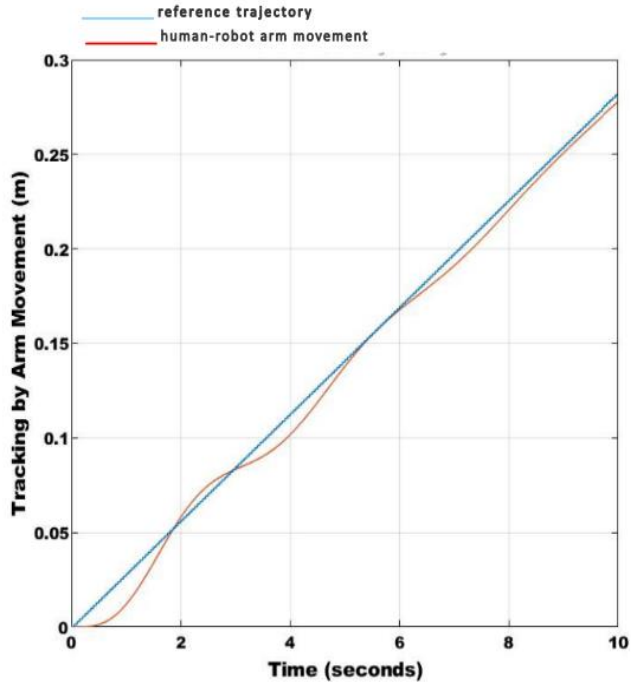

Figure 15: Tracking performance of plant under the influence of the learning constraint controller and both feedforward P-ILC and feedback PD controller

## 5 DISCUSSIONS

**Scenario 1: Effect of PD controller on Plant's tracking performance.**

In this instance, the feedback controller acts alone in improving the tracking performance of the plant. Under a simulation period of 10 seconds, it was observed after 16 trials that the tracking error was non-changing and constant Fig. 9. Using equations 23 and 24, a constant displacement error of 0.0168 m (5.94%) with respect to the trajectory range was obtained when the PD controller acted alone. Table (2) shows the non-changing performance at several trials.

**Scenario 2: Effect of combined feedback and feedforward controller on plant's tracking performance.**

In this section, the influence of the feed-forward controller on plant's tracking performance is evaluated over multiple iterations and four different learning gain space "L".

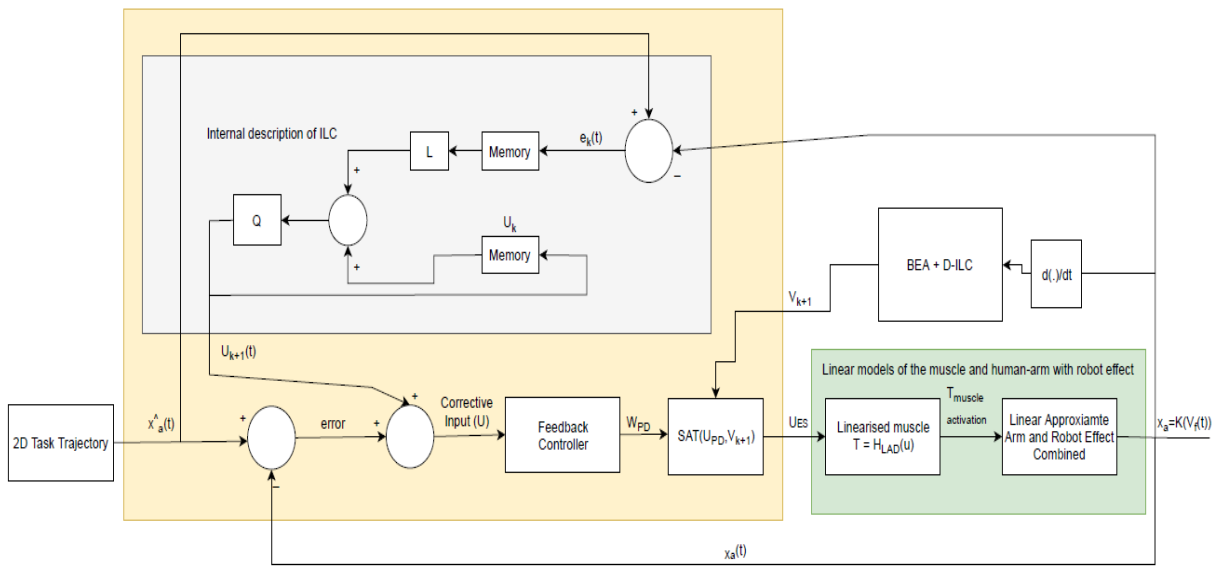

Figure 16: Block diagram of the implemented P-ILC + feedback + D-ILC based BEA-SAT controller.

At L=0.1, the plant displacement error during trajectory tracking improved from an error of 0.0168 m (5.94%) at the first iteration to a steady value of 0.0066 m (2.33%) at the fifteen and sixteen iterations Fig. 10. At L=0.2, a faster learning transient was observed under eight iterations. The tracking displacement error improved from 0.0168m (5.94%) to 0.0066 (2.33%) at the eighth trial Table (3). At L= 0.8 and L = 0.9, the impact of the feed-forward control inputs on the plant tracking performance was very swift. Under two iterations, displacement error was minimized to 0.0064 m and 0.0060 m respectively Fig. 11.

**Scenario 3: Effect of learning constrain velocity controller on the plant's movement rate when combined with Feed-forward and Feedback controller.**

Using Fig. 16, the efficacy of the learning velocity constraint controller in ensuring that the transient error growth of P-ILC doesn't cause the PD controller to apply control inputs that can cause the plant model to go beyond its specific velocity is discussed.

Up to this point, scenario 2 has successfully proven that ILC aids improvement in plant displacement errors when tracking a given task-plane trajectory. The specific velocity $\dot{r}_h$ for the plant model was identified as 0.4791 m/s in absence of the P-ILC controller in scenario 1 and its corresponding PD energy component $V_{k=0}$ was evaluated as 0.4615.

At the first iteration, upon selection of the lowest P-ILC gain value L=0.1, a SAT constraining value $V_k = 0.4603$ was computed by the D-ILC-BEA algorithm using the velocity errors $\dot{e}_{k=0}$ and the obtained offset value $V_{k=0}$. When simulated over 10 seconds, the computed $V_k$ was applied via the SAT function on the growing PD output energy to obtain a non-violating velocity $\dot{r}_h$ of 0.4546 m/s in Fig. 12a and b. Nonetheless, Fig. 13 and 14b shows that D-ILC+BEA+SAT kept limiting the PD energy level due to P-ILC transient learning error from trial-to-trial Table (4).

For the second iteration, a SAT constraint value of $V_{k+1} = 0.4605$ was computed by D-ILC-BEA in Eqn. 15 and 17 using the $\dot{e}_{k=1}$ and $V_{k=1}$. Consequently, during trial, the PD energy signal grew to 0.5047 due to P-ILC transient error growth but was constrained by $V_{k+1}$. Over the next 12 iterations, the plant achieved a good tracking performance shown in Fig. 15 without violating the set plant-specific velocity illustrated in Fig. 14a. Because safety is paramount in rehabilitation, it's believed that the learning D-ILC + BEA algorithm ensured velocity isn't violated from trial to trial while producing a satisfactory tracking performance in Fig. 15.

## 7 CONCLUSIONS

This study as proved that ILC can be combined with other techniques to serve as constraint controller. P-ILC + PD improved tracking performance of the plant model over several iterations whilst P-ILC + PD + D-ILC+BEA guaranteed both good tracking performance and bounded patient-specific velocity.

So far, with the developed algorithms proving feasible, an experimental validation will be conducted with the PULSR hybrid robotic system at Obafemi Awolowo University Teaching Hospital Centre (OAUTHC).

## 7 ACKNOWLEDGEMENTS

This research was sponsored by a TETFUND National Research Fund grant.

## 8 CONFLICTS OF INTERESTS

None

## 9 REFERENCES

[1]. Markus H. Stroke: causes and clinical features. Medicine. 36(11), 586-91, 2008.

[2]. Muir KW. Stroke. Medicine (Abingdon). 41(3), 169–74, 2013.

[3]. Liu Y, Hong Y, Ji L. Dynamic Analysis of the Abnormal Isometric Strength Movement Pattern between Shoulder and Elbow Joint in Patients with Hemiplegia. Journal of healthcare engineering. 2018.

[4]. Zavoreo I, BAŠIĆ-KES VA, Demarin V. Stroke and neuroplasticity. Periodicum biologorum. 114(3), 393-6, 2012.

[5]. Li S. Spasticity, motor recovery, and neural plasticity after stroke. Frontiers in neurology. 8, 120, 2017.

[6]. Oña ED, Garcia-Haro JM, Jardón A, Balaguer C. Robotics in health care: Perspectives of robot-aided interventions in clinical practice for rehabilitation of upper limbs. Applied sciences. 9(13), 2586, 2019.

[7]. Cuesta-Gómez A, Molina-Rueda F, Carratala-Tejada M, Imatz-Ojanguren E, Torricelli D, Miangolarra-Page JC. The use of functional electrical stimulation on the upper limb and interscapular muscles of patients with stroke for the improvement of reaching movements: A feasibility study. Frontiers in neurology. 8, 186, 2017.

[8]. Mazzoleni S, Duret C, Grosmaire AG, Battini E. Combining upper limb robotic rehabilitation with other therapeutic approaches after stroke: current status, rationale, and challenges. BioMed research international. 2017.

[9]. Schauer T, Vrontos A. Modeling of mixed artificially and voluntary induced muscle contractions for controlled functional electrical stimulation of shoulder abduction. IFAC-PapersOnLine. 51(34), 284-9, 2019.

[10]. Alon G. Functional Electrical Stimulation (FES): Clinical successes and failures to date. Journal of Novel Physiotherapy and Rehabilitation. 2, 080-6, 2018.

[11]. Cheung VC, Niu CM, Li S, Xie Q, Lan N. A novel FES strategy for poststroke rehabilitation based on the natural organization of neuromuscular control. IEEE reviews in biomedical engineering. 12, 154-67, 2018.

[12]. Freeman CT, Rogers E, Burridge JH, Meadmore KL. Iterative learning control in health care: Electrical Stimulation and Robotic-Assisted Upper-Limb Stroke Rehabilitation. IEEE Control Systems Magazine. 32(1), 18 – 43, 2012.

[13]. Meadmore KL, Exell TA, Hallewell E, Hughes AM, Freeman CT, Kutlu M, Benson V, Rogers E, Burridge JH. The application of precisely controlled functional electrical stimulation to the shoulder, elbow and wrist for upper limb stroke rehabilitation: a feasibility study. Journal of neuroengineering and rehabilitation. 11(1), 1-1, 2014.

[14]. Freeman CT, Hughes AM, Burridge JH, Chappell PH, Lewin PL, Rogers E. A robotic workstation for stroke rehabilitation of the upper extremity using FES. Medical engineering & physics. 31(3), 364-73, 2009.

[15]. Kutlu M, Freeman CT, Hallewell E, Hughes A, Laila DS. FES-based upper-limb stroke rehabilitation with advanced sensing. IEEE International Conference on Rehabilitation Robotics

[16]. Sa-e S, Freeman CT, Yang K. Iterative learning control of functional electrical stimulation in the presence of voluntary user effort. Control Engineering Practice.96, 104303, 2020.

[17]. Sebastian G, Tan Y, Oetomo D, Mareels I. Input and Output Constraints in Iterative learning Control Design for Robotic Manipulators. Unmanned Systems. 6, 197-208, 2018.

[18]. Longman RW. Iterative learning control and repetitive control for engineering practice. International journal of control. 73(10), 930-54, 2000.

[19]. Yovchev K, Delchev K, Krastev E. Constrained output iterative learning control. Archives of Control Sciences. 30, 2020.

[20]. Ayodele KP, Akinniyi OT, Oluwatope AO, Jubril AM, Ogundele AO, Komolafe MA. A Simulator for Testing Planar Upper Extremity Rehabilitation Robot Control Algorithms. Nigerian Journal of Technology. 40(1), 115-28, 2021.

[21]. Song P, Yu Y, Zhang X. A tutorial survey and comparison of impedance control on robotic manipulation. Robotica. 37(5), 801-36, 2019.

[22]. Durfee WK, Palmer KI. Estimation of force-activation, force-length, and force-velocity properties in isolated, electrically stimulated muscle. IEEE Transactions on Biomedical Engineering. 41(3), 205-16, 1994.

[23]. Le F, Markovsky I, Freeman CT, & Rogers E. Identification of electrically stimulated muscle models of stroke patients. Control Engineering Practice. 18(4), 396–407, 2010.

[24]. Houda B, Nahla K, Safya B. Musculoskeletal Modeling of Elbow Joint under Functional Electrical Stimulation," International Conference on Advanced Systems and Emergent Technologies (IC_ASET). 307-310, 2019.

[25]. Fadali MS, Visioli A. Digital control engineering. Analysis and design. Elsevier, Waltham, pp. 157-165, 2013.

[26]. Moore KL, Chen Y, Ahn HS. Iterative learning control: A tutorial and big picture view. In Proceedings of the IEEE Conference on Decision and Control. 45, 2352-2357, 2006.

[27]. Taragna M, Cannizzaro D, Votano S. Vibration compensation for robotic manipulators by iterative learning control. POLITECNICO DI TORINO. 2018.

[28]. Mahamood RM. Adaptive Controller Design for Two-Link Flexible Manipulator. InIAENG Transactions on Engineering Technologies. Springer, 115-128, 2014.

[29]. Yovchev K, Delchev K, Krastev E. State space constrained iterative learning control for robotic manipulators. Asian Journal of Control. 20(3), 1145-50, 2018.

[30]. Yovchev K, Delchev K, Krastev E. Constrained Output Iterative Learning Control. Archives of Control Science.30(66), 157-176, 2020.

[31]. Youngsma, K. Development of a Model and Simulation Framework for a Modular Robotic Leg. Worcester Polytechnic Institute, 2012.

[32]. Ardakani MG, Khong SZ, Bernhardsson B. On the convergence of iterative learning control. Automatica. 78, 266 – 273, 2017.